\documentclass[10pt,twocolumn,letterpaper]{article}

\usepackage{iccv}
\usepackage{times}
\usepackage{epsfig}
\usepackage{graphicx}
\usepackage{amsmath}
\usepackage{amssymb}
\usepackage[dvipsnames]{xcolor}

\usepackage[accsupp]{axessibility} %

\usepackage{graphicx}
\usepackage{amsmath}
\usepackage{amssymb}
\usepackage{booktabs}
\usepackage{overpic}

\usepackage{multirow}
\usepackage{makecell}
\usepackage{colortbl}  %
\usepackage[pagebackref=true,breaklinks=true,letterpaper=true,colorlinks,citecolor=NavyBlue,bookmarks=false]{hyperref}

\newcommand{\yu}[1]{\textcolor[rgb]{0, 0, 0}{#1}}
\newcommand{\kk}[1]{\textcolor[rgb]{0,0,0}{#1}}
\newcommand{\fang}[1]{\textcolor[rgb]{0, 0, 0}{#1}}
\newcommand{\xk}[1]{\textcolor[rgb]{0,0,0}{#1}}
\newcommand{\liu}[1]{\textcolor[rgb]{0., 0., 0.}{#1}}

\newcommand{\ryn}[1]{\textcolor[rgb]{0, 0, 0}{#1}}
\newcommand{\liufang}[1]{\textcolor[rgb]{0,0,0}{#1}}

\definecolor{mygray}{gray}{.9}

\iccvfinalcopy %

\def\iccvPaperID{4825} %

\ificcvfinal\fi

\begin{document}

\title{Referring Image Segmentation Using Text Supervision}

\author{Fang Liu$^{1,2,}$\footnotemark[1] , Yuhao Liu$^{2,}$\footnotemark[1] , Yuqiu Kong$^{1,}$\footnotemark[2]  , Ke Xu$^{2,}$\footnotemark[2] , Lihe Zhang$^{1}$ , \\ 
Baocai Yin$^{1}$, Gerhard Hancke$^{2}$ , Rynson Lau$^{2,}$\footnotemark[2]\\
$^{1}$ Dalian University of Technology,  $^{2}$ City University of Hong Kong\\
{\tt\small \{fawnliu2333, yuhaoLiu7456, kkangwing\}@gmail.com,}  
{\tt\small\{yqkong, zhanglihe, ybc\}@dlut.edu.cn, }\\
{\tt\small\{gp.hancke, rynson.lau\}@cityu.edu.hk}
\and
}

\maketitle
\ificcvfinal\thispagestyle{empty}\fi

\footnotetext[1]{Joint first authors. $^{\dagger}$Joint corresponding authors.
}

\begin{abstract}
    \xk{Existing Referring Image Segmentation (RIS) methods typically require expensive pixel-level or box-level annotations for supervision.} \ryn{In this paper, we observe that the referring texts used in RIS already provide sufficient information to localize the target object. Hence, we propose a novel weakly-supervised RIS framework to formulate the target localization problem as a classification process to differentiate between positive and negative text expressions. While the referring text expressions for an image are used as positive expressions, the referring text expressions from other images can be used as negative expressions for this image.}
    Our framework has three \ryn{main} novelties.
    First, we propose a bilateral prompt method to facilitate the classification process, by harmonizing the domain discrepancy between visual and linguistic features.
    Second, we propose a calibration method to reduce noisy background information and improve the \ryn{correctness of the response maps} for  target object localization.
    Third, \ryn{we propose a positive response map selection strategy to generate high-quality pseudo-labels from the enhanced response maps, for training a segmentation network for RIS inference}.
    \xk{For evaluation, we propose a new metric to measure localization accuracy.}
    Experiments on four benchmarks show that our framework achieves promising performances to existing fully-supervised RIS methods while outperforming state-of-the-art weakly-supervised methods adapted from related areas. 
    Code is available at \href{https://github.com/fawnliu/TRIS}{https://github.com/fawnliu/TRIS}.
    \vspace{-5mm}
\end{abstract}

\section{Introduction}
\label{sec:intro}
    Referring Image Segmentation (RIS) aims to segment \ryn{a target} object from an input image according to the input linguistic query.
    It has many applications such as text-based image editing~\cite{chen2018language, Choi_2022_ECCV}, human-computer interaction~\cite{wang2019reinforced,actM}, E-commercial search engine~\cite{zhuge2021kaleido,Zheng2023MAKEVP}.
    
\begin{figure}[htp]
\begin{center}
\begin{tabular}{c@{\hspace{0.8mm}}c@{\hspace{0.8mm}}c@{\hspace{0.8mm}}c}
\put(-40,0){\footnotesize Q: \textit{``a woman wearing the cream color dress and cutting cake with a man"}} \\ 
\includegraphics[width=0.31\linewidth,height=0.21\linewidth]{figs/teasor.png}&
\includegraphics[width=0.31\linewidth,height=0.21\linewidth]{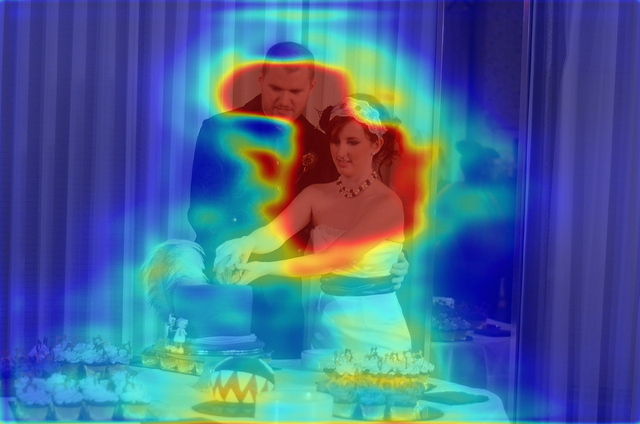}&
\includegraphics[width=0.31\linewidth,height=0.21\linewidth]{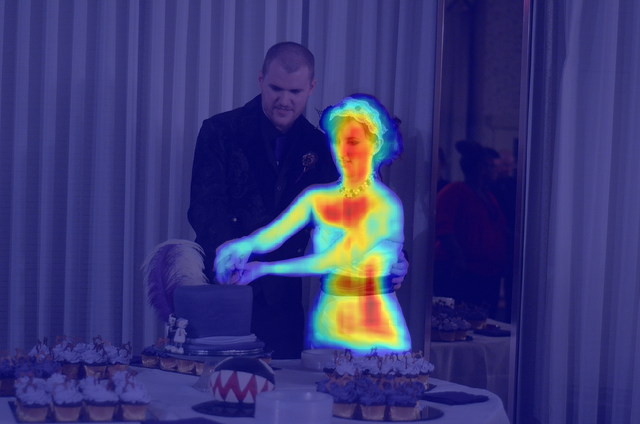}& \\
\fontsize{8pt}{\baselineskip}\selectfont (a) Image &
\fontsize{8pt}{\baselineskip}\selectfont  (b) WWbL \cite{shaharabany2022} &
\fontsize{8pt}{\baselineskip}\selectfont  (c) LAVT \cite{yang2022lavt}\vspace{1mm}\\
\includegraphics[width=0.31\linewidth,height=0.21\linewidth]{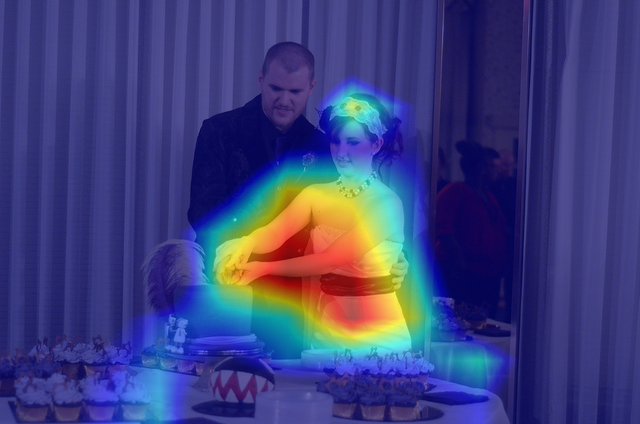}&
\includegraphics[width=0.31\linewidth,height=0.21\linewidth]{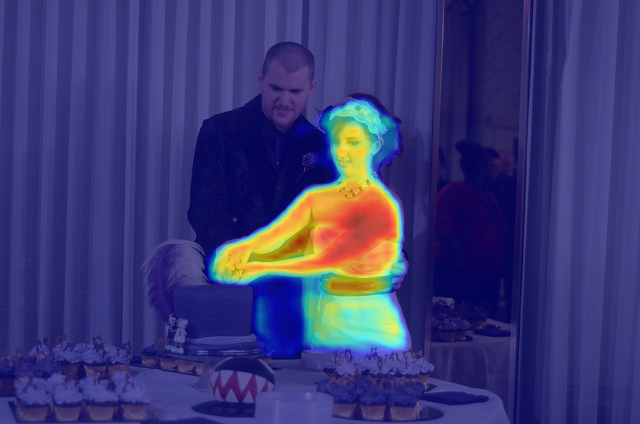}&
\includegraphics[width=0.31\linewidth,height=0.21\linewidth]{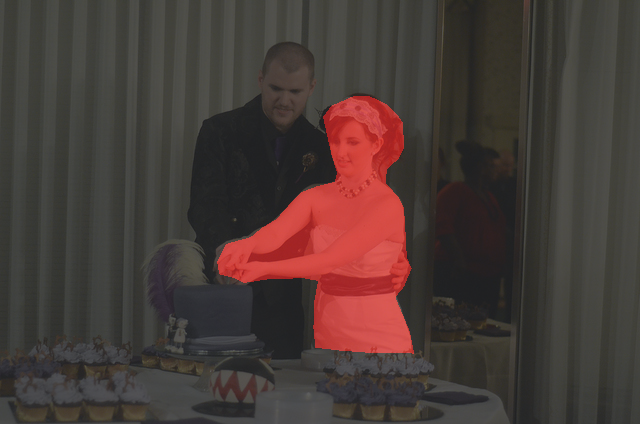}& \\
\fontsize{8pt}{\baselineskip}\selectfont (d) Ours (Step-1) &
\fontsize{8pt}{\baselineskip}\selectfont  (e) Ours (Step-2) &
\fontsize{8pt}{\baselineskip}\selectfont  (f) GT \\
\end{tabular}
\end{center}
\vspace{-3mm}
\caption{Given an input image and a query expression (a), our method leverages a text-to-image response optimization process to locate the target object in the first step (d), and then produce the final map using a segmentation network in the second step (e). Our method outperforms the weakly-supervised method~\cite{shaharabany2022} (b) and achieves \liu{competitive} result compared to the state-of-the-art fully-supervised RIS method~\cite{yang2022lavt} (c). Note that our method is trained using only text expressions\ryn{, which are already available}.}
\label{fig:teasor}
\vspace{-4mm}
\end{figure}

    Hu~\etal~\cite{hu2016segmentation} propose the first RIS method, which separately extracts visual and linguistic features \ryn{and} then concatenates \liu{them} to predict the segmentation map.
    Subsequently, many methods are proposed to fuse multi-modal features for RIS in different ways,
    \eg, with unidirectional fusion~\cite{ye2019cross, liu2021cross} and bidirectional fusion~\cite{hu2020bi,feng2021encoder} using CNNs, and long-range cross-modal modeling~\cite{yang2022lavt,kim2022restr} using transformers.
    \kk{However, almost all existing RIS methods rely on the fully-supervised learning scheme to produce accurate results, which requires time-consuming and labour-intensive pixel-level annotations}. \ryn{(For example, skilled annotators may need an average of $79s$ \cite{papadopoulos2017extreme} to label a polygon-based instance mask for an image in MS COCO \cite{lin2014microsoft}.)}
    \kk{Recently, Feng~\etal~\cite{feng2022learning} propose to use bounding box annotations for weakly-supervised RIS, but it is still costly for labeling large-scale RIS datasets (on average 7$s$ to annotate a single bounding box).}
    Hence, there is still a need to design RIS models with cheaper supervision signals.

    Besides using bounding boxes, there are also other kinds of weak supervision signals studied in other computer vision tasks, including scribbles~\cite{zhang2020weakly,yu2021structure}, points~\cite{gao2022weakly} and class labels~\cite{wang2017learning, piao2021mfnet}.
    However, scribbles and points still require and are sensitive to human involvement. Although class labels work well for other visual tasks, \eg, salient object detection~\cite{wang2017learning}, they are unsuitable for RIS due to their lack of instance information.

    \xk{We observe that the input \liu{referring} texts used in \ryn{the RIS task} have already provided distinctive descriptions\ryn{, which describe one or several properties, of the target object that can be used to locate the object}. %
    Inspired by this, we propose a novel weakly-supervised RIS framework, which \ryn{learns to locate the target objects by learning how to classify positive and negative texts, where the positive texts are the referring texts that are used to describe the corresponding \liu{objects of the input} images while the negative texts are the referring texts that describe objects from other images.}}
    {Our framework has three main technical novelties.}
    {First,} to facilitate classification process, we propose a bilateral prompt method to harmonize the modal discrepancy between visual and linguistic features.
    {Second}, we propose a calibration method \ryn{that consists of a foreground enhancement process and a background suppression process} to enhance the derived response maps for target object localization.
    Third, we propose a novel positive response map selection strategy to derive high-quality pseudo-labels from the enhanced response maps, which are used to train \ryn{a} segmentation network for RIS inference. \xk{In addition, we propose a new metric to evaluate localization accuracy, which can reduce  \liu{in-box} errors of the existing pointing-game accuracy metric.}
    
    As shown in Fig.~\ref{fig:teasor}, our method can locate the target object and produce the segmentation map \yu{accurately}. \ryn{Even though it is trained using only text} expressions, it \ryn{produces a comparable}
    result to the fully-supervised method~\cite{yang2022lavt}. 
    In \ryn{summary}, this paper has the following main contributions:
    \begin{itemize}
    \item \xk{We propose a novel \liu{weakly-supervised} RIS framework} that \ryn{is supervised only by the readily available referring texts} and does not require any extra annotations.
    
    \item \xk{Our framework has three main} technical novelties. First, we propose a bilateral prompt method to harmonize the visual and linguistic modal discrepancy. Second, we propose a calibration method to improve the correctness of the response maps for localization. Third, we propose a response map selection strategy to generate high-quality pseudo labels for the segmentation of the target objects.

    \item \xk{We propose a new metric for the evaluation of localization accuracy.} Extensive experiments on four benchmarks show that our framework can produce promising results compared to previous fully-supervised RIS methods and outperforms existing weakly-supervised baselines adapted from related tasks.
    
    \end{itemize}
\begin{figure*}[t!]
  \centering
   \includegraphics[width=1\linewidth]{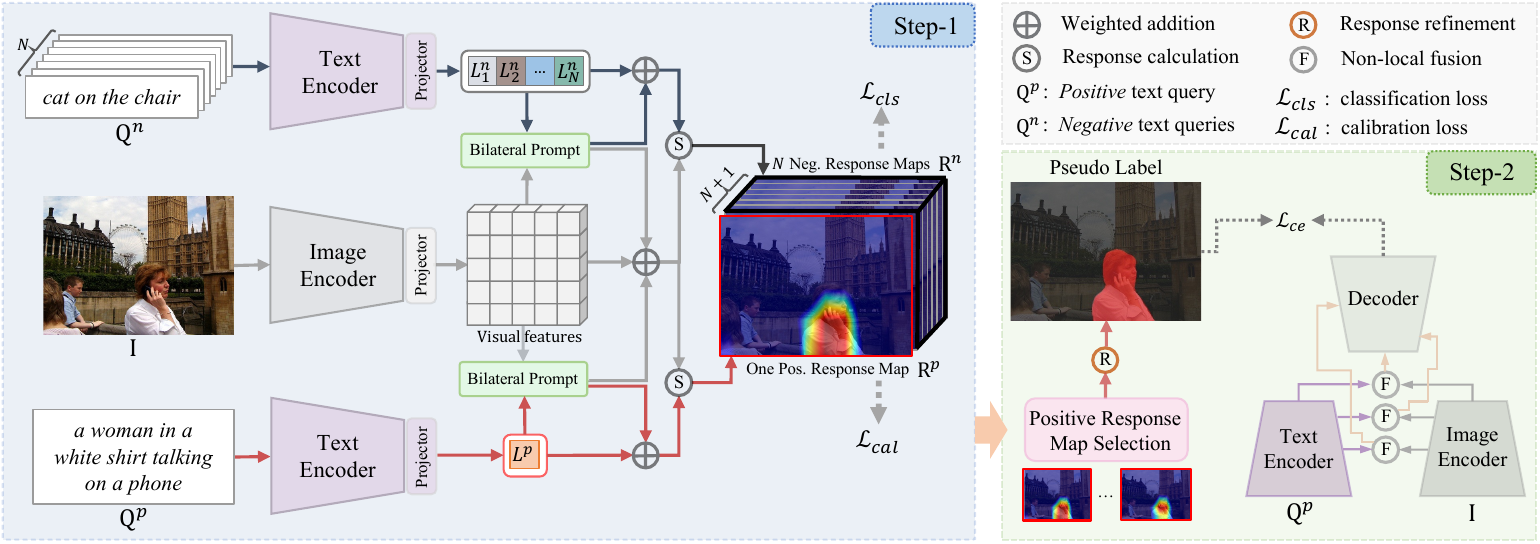}
   \vspace{-5mm} 
   \caption{Our weakly-supervised RIS framework \ryn{has two steps. \textbf{Step-1}: It learns to classify input text expressions, and uses the \ryn{positive text} expression to help localize the target \ryn{object to produce} response maps. \textbf{Step-2}: It feeds the pseudo-labels derived from the response maps to train a segmentation network for RIS inference.} We propose a bilateral prompt method to harmonize the multi-modal discrepancy and a calibration method to enhance the response maps in \ryn{Step-1}. \ryn{We propose a positive response map selection strategy in Step-2 to help select the best response maps as pseudo-labels}. }
   \label{fig:pipeline}
\end{figure*}

\section{Related Works}
\label{sec:related}
    {\flushleft \bf Referring Image Segmentation} detects and segments target objects according to the input text expressions. Hu~\etal~\cite{hu2016segmentation} propose the first RIS method with a CNN model. \ryn{Many follow-up methods are then proposed, mainly focusing on \ryn{addressing} the linguistic and visual feature fusion problem}.
    Some methods adopt \fang{recurrent refinement \cite{chen2019see, li2018referring, liu2017recurrent}} or dynamic filters~\cite{margffoy2018dynamic,chen2019referring,jing2021locate} to fuse visual and linguistic features.
    To capture long-range dependencies between two modalities, several methods explore \fang{attention mechanisms~\cite{ye2019cross,hu2020bi,feng2021encoder, liu2021cross, shi2018key, luo2020mcn, yu2018mattnet, wu2020phrasecut, hui2020linguistic, yang2021bottom}} or transformer-based architectures~\cite{ding2021vln, li2021reftr, kim2022restr, yang2022lavt, wang2022cris, zhu2022seqtr, liu2023local}.
    
    However, all these methods are fully-supervised. Training them requires \ryn{pixel-level annotations}, which are time-consuming and labor-intensive to obtain.
    Recently, Feng~\etal~\cite{feng2022learning} propose the first weakly-supervised RIS method, \ryn{which is based on bounding box annotations. Although bounding boxes are cheaper to annotate, they still require a significant amount of effort on a large-scale dataset.}
    In this paper, we propose \ryn{a weakly-supervised method that uses only the available text expressions for training}.
    
    {\flushleft \bf Weakly-Supervised Learning} \ryn{tries to reduce labelling efforts with different kinds of weak labels}, \eg, bounding boxes~\cite{dai2015boxsup,liang2022weakly},    scribbles~\cite{zhang2020weakly,yu2021structure}, points~\cite{gao2022weakly}, class labels~\cite{zhou2016learning, pathak2014fully, wang2017learning, araslanov2020single, xie2022clims, tian2022learning, tian2020weakly},
    and text expressions~\cite{shaharabany2022, datta2019align2ground, gupta2020contrastive, liu2021relation, wang2021improving, jiang2022pseudo, akbari2019multi, arbelle2021detector}, \ryn{for various} computer vision tasks.
    
    Among \ryn{these proposed weak labels, image-level class labels are very popular, as they provide} high-level semantic information but are relatively cheap to label. 
    The above methods propose different class activation map generation processes to locate the objects, \eg, global average pooling~\cite{zhou2016learning}, global max pooling~\cite{pathak2014fully}, global smoothing pooling~\cite{wang2017learning}, and normalized global weighted pooling~\cite{araslanov2020single}. 
    Xie \etal \cite{xie2022clims} propose to further use background class labels to improve the quality of maps. 
    \ryn{While also using text input}, class \ryn{labeling does not work in} our task due to the  lack of instance-level information.
    
    Language expressions are commonly used as weak supervision signals in the weakly-supervised visual grounding (WSG) task
     (\ie, detecting the target objects with bounding boxes according to the input expressions). 
     However, most WSG methods~\cite{datta2019align2ground, gupta2020contrastive, liu2021relation, wang2021improving, jiang2022pseudo, wang2019phrase} resort to additional pre-trained object detectors to produce a set of \liu{proposals} and then select the one with the highest confidence score by matching visual features with phrase features. 
    To avoid the dependence on pre-trained detectors, Arbelle~\etal~\cite{arbelle2021detector} propose a self-supervised method to train the network by randomly blending the images according to the expressions. 
    \liu{Shaharabany~\etal~\cite{shaharabany2022} compute the relevancy heatmaps and treat them as GTs to supervise the generated response maps.}
    Akbari~\etal~\cite{akbari2019multi} propose a posterior probability-based multi-modal loss to guide the network to predict correlation scores for the related image-sentence pairs.
    
    While these methods are close in spirit to our work, they only detect the target objects with bounding boxes. \ryn{Our experiments demonstrate} that it is non-trivial to obtain pixel-wise segmentation maps from their bounding-box results. \ryn{In contrast, our} method learns to detect the target objects in a pixel-wise manner \ryn{via a} text-to-image optimization process, which allows \fang{accurate localization \ryn{of} the target}.

\section{Our Framework}
\label{sec:method}

The main objective of the weakly-supervised RIS task is to establish a pixel-wise association between the visual content and the input referring expressions without using \ryn{pixel-level} annotations.
\liu{We note that the input referring expressions inherently possess discriminative information pertaining to the localization of the target objects or regions. 
In light of this observation, our framework \ryn{is designed to learn to classify positive and negative expressions of each input image, through which it also learns to localize the target object in the image as described by the positive expressions. 
The positive expressions are the referring text expressions that are used to describe the target object of the input image, while the negative expressions are the referring text expressions taken from other images.}
\ryn{Our} classification process involves the modelling of text-to-image responses, which can \ryn{learn to associate the visual contents in the input image with the positive} expressions.
}

Fig.~\ref{fig:pipeline} shows our weakly-supervised RIS framework, which has two steps. The first step models the text-to-image response of the classification process to locate the target \ryn{object as specified by the text} expressions.
We propose a bilateral prompt method in this step to harmonize the discrepancy \ryn{between the} visual and linguistic features, and a calibration method to enhance the completeness and correctness of the \liu{positive} response maps.
\ryn{The second step leverages the response maps from the first step to produce pseudo labels. These pseudo labels are then used to train a segmentation network for RIS inference.}
\yu{A positive response map selection strategy is proposed to select the best response \ryn{maps} for pseudo-label generation.}

\subsection{\ryn{Text-to-Image Response Modeling}}
\label{subsec:overview}

We first explain how we \ryn{model text-to-image} responses, which are used to localize the target objects \ryn{as specified by the} expressions in the query classification process.

Given an input image $\mathbf{I}\in \mathbb{R}^{H_I\times W_I \times 3}$ and a text expression query $\mathbf{Q}\in\mathbb{R}^{T}$ \liu{with $T$ words}, 
we first extract the visual features \liufang{$\mathbf{V}_e \in \mathbb{R}^{H \times W \times C_{v}}$} and text features \liufang{$\mathbf{L}_e \in \mathbb{R}^{1\times C_{l}}$} by an image encoder ${E}_{v}$ and a text encoder ${E}_{l}$ \cite{radford2021clip}, where $H=H_I/s$ and $W=W_I/s$.
$C_{v}$ and $C_{l}$ denote the numbers of channels of visual and text features, respectively. $s$ is the ratio of down-sampling.
We then use a projector layer to transform $\mathbf{V}$ and $\mathbf{L}$ to \ryn{a} unified hidden dimension $C_{d}$, \ie, $\mathbf{V} \in \mathbb{R}^{ H \times W  \times C_{d}}$ and $\mathbf{L} \in \mathbb{R}^{1 \times C_{d}}$. 
The $L_2$ channel-wise normalization is used to regularize the output of the projection layer.
To harmonize the domain discrepancy, we propose a bilateral prompt method (Sec.~\ref{subsec:prompt}) to update the visual and text features as:
\begin{align} \label{eq:prompt}
\hat{\mathbf{V}} = \mathbf{V} + \alpha  \mathbf{V}^{'}; \quad 
    \hat{\mathbf{L}} = \mathbf{L} + \beta  \mathbf{L}^{'} , \nonumber \\ 
    \text{with}~~~~\mathbf{V}^{'}, \mathbf{L}^{'} = \varGamma_\text{Prompt}\left(\mathbf{V}, \mathbf{L}\right) ,
\end{align}
where \ryn{$\hat{\mathbf{V}}$ and $\hat{\mathbf{L}}$ denote the updated visual and text features, and $\mathbf{V}^{'}$ and $\mathbf{L}^{'}$ are the residual \yu{enrichment prompted}  from $\mathbf{L}$ and $\mathbf{V}$}. $\alpha$ and $\beta$ are  weights to control \yu{the proportion of} bilateral prompting.
\liu{$\varGamma_\text{Prompt}$ is \ryn{our} bilateral prompt method.}

We \ryn{have now aligned} $\hat{\mathbf{V}}$ and $\hat{\mathbf{L}}$ explicitly. 
\liu{After reshaping, the response between $i$-th flattened pixel of $\hat{\mathbf{V}}$ and $j$-th query of $\hat{\mathbf{L}}$} can then be calculated as:
\begin{align} \label{eq:response}
    \mathbf{R}_{i,j}=  \sum_{\vartheta=0}^{C_{d}} \hat{\mathbf{V}}_{i, \vartheta} \cdot \hat{\mathbf{L}}_{j, \vartheta},
\end{align}
where $\vartheta$ and $\cdot$ represent channel index and element-wise multiplication, respectively. 
\liu{We employ a learnable temperature parameter $\tau$~\cite{radford2021clip} to scale and constrain the range of $\mathbf{R}_{i,j}$.}
In this way, we establish the \ryn{relationship} of each pixel to the whole referring expression $\mathbf{Q}$, and different pixels will echo responses of different degrees. A higher response indicates that the pixel is more likely to belong to the target \ryn{object} \kk{described by the \ryn{positive} expression.}

\subsection{Bilateral Prompt}
\label{subsec:prompt}

\liu{To facilitate learning knowledge from the pretrained model \cite{radford2021clip}}, we propose a bilateral prompt method to harmonize the domain discrepancy between visual and text features. 
Unlike previous \fang{prompt} methods that either refine visual and text features separately~\cite{gao2021clip}, or \ryn{use the visual features to refine the} text features~\cite{rao2022denseclip, zhou2022learning, zhou2022conditional}, our bilateral prompt method enhances the features of one modality with those of the other modality. 
Enriching the text \ryn{features with the} visual features facilitates the classification process, while enriching the visual \ryn{features with the} text features helps \ryn{localize the} target objects.

\begin{figure}[t]
  \centering
   \includegraphics[width=0.96\linewidth]{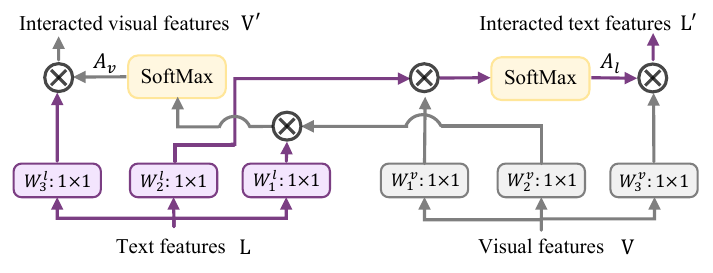}
   \vspace{-3mm} 
   \caption{Our bilateral prompt method
   \ryn{aims to reduce} the modality differences between visual and linguistic features. The text \ryn{features $\mathbf{L}$ are enhanced by the visual features} $\mathbf{V}$ via the attention map $A_l$, which updates $\mathbf{L}$ to $\mathbf{L^{'}}$ for classification. The visual \ryn{features $\mathbf{V}$ are enhanced by the text features} $\mathbf{L}$ via the attention map $A_v$ to update $\mathbf{V}$ to $\mathbf{V^{'}}$ for localization.}
   \label{fig:prompt}
\end{figure}

Fig.~\ref{fig:prompt} shows our bilateral prompt method. Given input features $\mathbf{V} \in \mathbb{R}^{H \times W \times C_{d}}$ and $\mathbf{L} \in \mathbb{R}^{1\times C_{d}}$, we first compute two attention maps as:
\begin{align}  \label{eq:att1}
    & \mathbf{A}_{l} = {\rm SoftMax} \left( (\mathbf{V} W^{{v}}_{1}) \otimes
    (\mathbf{L} W^{{l}}_{2})^\top/\sqrt{C_{{d}}} \right) , \\
    & \mathbf{A}_{v} = {\rm SoftMax} \left( (\mathbf{L} W^{{l}}_{1}) \otimes (\mathbf{V} W^{{v}}_{2})^\top/\sqrt{C_{{d}}} \right)  , 
\end{align}
where \yu{$\mathbf{A}_{l}\in \mathbb{R}^{1\times HW}$ and $\mathbf{A}_{v}\in\mathbb{R}^{HW \times 1}$ denote the affinity propagation of visual-to-text features and text-to-visual features, respectively}. 
\liufang{$W^{v}_*\in\mathbb{R}^{C_d\times C_d}$ and $W^{l}_*\in\mathbb{R}^{C_d\times C_d}$} are the learnable parameters for $\mathbf{V}$ and $\mathbf{L}$. 
\liu{$\otimes$ denotes matrix multiplication.}
The \ryn{bilateral prompt} is then formulated as:
\begin{align} \label{eq:prompt2}
    \mathbf{L}^{'} = \mathbf{A}_{l}^\top \otimes (\mathbf{V} W_{3}^{v}), 
    \quad 
    \mathbf{V}^{'} = Re(\mathbf{A}_{v}^\top \otimes (\mathbf{L} W_{3}^{l})),
\end{align}
where \yu{$Re$ is a shape transform function},  $\mathbf{L}^{'}\in\mathbb{R}^{1\times C_{d}}$ and
 $\mathbf{V}^{'}\in \mathbb{R}^{H \times W \times C_{d}}$ \liu{denote visual-enhanced linguistic features and language-guided visual features, respectively}. 
In this way, the learned bilateral residuals \liufang{(\ie, $\mathbf{L}^{'}$ and $\mathbf{V}^{'}$)} can be used to update both visual and linguistic features to effectively harmonize their discrepancy, \liufang{according to Eq. \eqref{eq:prompt}.}

\subsection{Localization via \ryn{Text-to-Image} \yu{Optimization}}
\label{subsec:optim}

{\flushleft \bf Classification.} We formulate a classification process for the network to \ryn{learn to select the \fang{positive} expressions} 
from a set of \ryn{positive and negative} expressions, \ryn{with} which we optimize the text-to-image response maps to localize the target object. 
We implement the classification process in a contrastive learning manner~\cite{he2020momentum}.
In our problem, we use the input image $\mathbf{I}$, \ryn{a positive text expression $\mathbf{Q}^{p} \in \mathbb{R}^{T}$ as a positive sample (or \fang{the anchor}), \yu{and} randomly sample $N$ \ryn{negative} text expressions $\mathbf{Q}^{n} \in \mathbb{R}^{N \times T}$ (\ie, text expressions of other images) \yu{from the whole dataset} as negative samples}.
\ryn{We} compute the response maps $\mathbf{R}^{a} \in \mathbb{R}^{H W \times (1+N)}$ \ryn{(which include $\mathbf{R}^{p} \in \mathbb{R}^{H W \times 1}$ for the positive expression $\mathbf{Q}^{p}$, and $\mathbf{R}^{n} \in \mathbb{R}^{H W \times N}$ for the negative expressions $\mathbf{Q}^{n}$) of image $\mathbf{I}$, as shown in Fig.~\ref{fig:pipeline}.}  
\ryn{We then} compute the image-level score $y_{j}$ for each \ryn{text} query $\mathbf{Q}_{j}$ \ryn{as}:
\begin{align} \label{eq:aggregation}
    y_{j} = \mathop{{\rm max}}_{i} \  \mathbf{R}^{a}_{i,j} + \frac{1}{HW} \sum_{i=1}^{HW} \mathbf{R}^{a}_{i,j} + \psi(\mathbf{R}^{a}_{\cdot,j}),
\end{align}
where $\psi(\mathbf{R}^{a}_{\cdot,j})$ is a regularization term proposed by~\cite{lin2017focal}. We use it to re-balance our negative and positive \ryn{text} queries. 
$\mathbf{y}_j$ ranges from 0 to 1, and the larger it is, the better the current query $\mathbf{Q}_{j}$ matches with the input image.

The classification process is then supervised by:
\begin{align} \label{eq:loss_cls}
    \mathcal{L}_{cls}(y, z) = -\frac{1}{N+1}&\sum_{j=1}^{1+N}  z_j   {\rm log}\left( \frac{1}{1+e^{-y_j}} \right) \nonumber \\
    &+ (1-z_j)  {\rm log} \left(\frac{e^{-y_j}}{1+e^{-y_j}} \right), 
\end{align}
where $\mathbf{z} \in \mathbb{R}^{1 \times (1+N)}$ is \ryn{an easy to obtain}  supervision signal, \ryn{with 1 indicating a positive query and 0 a negative one}.

{\flushleft \bf Calibration.} Given the coarse response \fang{maps} derived from the classification process, we propose a calibration method \yu{to enhance \kk{the correctness of the positive response map \yu{$\mathbf{R}^{p}$}} \ryn{by contrasting the target object with other objects of the same image (\ie, considering them as background noise).}}
\yu{Specifically, we multiply the input image $\mathbf{I}$ with $\mathbf{R}^{p}$ to obtain the target object and use it as the anchor.}
We use the \ryn{positive} query $\mathbf{Q}^{p}$ as the positive sample, and additionally sample $K$ queries that describe \ryn{other} objects of the same image as negative samples (\ie, $\mathbf{F}_{k}^{n}, k\in 1,2, \ldots, K$).
The calibration process can then be formulated as: 
\begin{align} \label{equ:cali}
    \mathcal{L}_{cal} = -({\rm log}\ S(\mathbf{I},\mathbf{R}^{p},\mathbf{Q}^{p}) + \sum_{k=1}^K {\rm log} \left(1-S(\mathbf{I},\mathbf{R}^{p},\mathbf{F}^{n}_{k}\right))), 
\end{align}
where $S(\cdot,\cdot,\cdot)$ is a  similarity function to measure the matching scores of the target object and a query, as:
\begin{align} \label{equ:score}
    S(\mathbf{I}, \mathbf{R}^{p}, \mathbf{Q}) = \varphi(E_{v}(\mathbf{I} 
\odot up(\mathbf{R}^{p})), E_{l}(\mathbf{Q})),
\end{align}
where $E_{v}$ and $E_{l}$ are CLIP~\cite{radford2021clip} visual and text encoders. \liufang{$\odot$ and $up()$ are Hadamard product and \ryn{an} up-sampling function, respectively.}  
$\varphi(\cdot,\cdot)$ \ryn{computes cosine similarity score}.

Eq.~\eqref{equ:cali} calibrates the positive response map in two \ryn{ways}. \yu{The first term helps learn more compact foreground regions by encouraging higher correlation of the instance-wise target object to the positive query.}
\fang{The second term suppresses noisy information of other background objects by decreasing the matching scores between foreground object regions and those negative queries.}

\begin{table*}[t!]
\setlength{\tabcolsep}{4pt}
\small
\centering
\renewcommand{\arraystretch}{1.0}
\begin{tabular}
{c|l|cc|cccccccccc}
\toprule
 \multirow{2}{*}{Metric} &\multirow{2}{*}{Method}&\multirow{2}{*}{$Sup.$}   &\multirow{2}{*}{Backbone}
 &
\multicolumn{1}{c}{ReferIt} 
&\multicolumn{3}{c}{RefCOCO}
&\multicolumn{3}{c}{RefCOCO+} 
&\multicolumn{3}{c}{RefCOCOg}  \\
&&& &test    &val&testA&testB   &val&testA&testB   &val (G)  &val (U) &test (U)\\
\midrule 
\multirow{9}{*}{IoU} &LAVT~\cite{yang2022lavt} &$\mathcal{F}$ &Swin-Base &-  &72.73 &75.82 &68.79 &62.14 &68.38 &55.10 &61.24 &62.09 &60.50 \\ 
\cline{2-14}
&CTS \cite{feng2022learning} &$\mathcal{B}$+$\mathcal{T}$ &ResNet-101  &62.44 &58.01 &60.52 &55.48 &47.12 &50.86 &40.26 &46.03 &- &- \\ 
\cline{2-14}
&\liu{AMR}$^\dagger$ \cite{qin2022activation}  &$\mathcal{T}$  &ResNet-50 &18.98 &14.12 &11.69 &17.47 &14.13 &11.47 &18.13 &15.83 &15.46 &15.59   \\ 
&\liu{GroupViT}$^\dagger$ \cite{xu2022groupvit} &$\mathcal{T}$ &GroupViT  &21.73 &18.03 &18.13 &19.33 &18.15 &17.65 &19.53 &19.97 &19.80 &20.09   \\ 
 &CLIP-ES$^\dagger$ \cite{lin2022clip} &$\mathcal{T}$ &ViT-Base &18.67  &13.79 &15.23 &12.87 &14.57 &16.01 &13.53 &14.16 &13.93 &14.09 \\ 
&$\text{GbS}^{\dagger}$ \cite{arbelle2021detector}  &$\mathcal{T}$ &VGG16  &14.21  &14.59  &14.60  &14.97  &14.49  &14.49  &15.77  &14.21  &13.75  &14.20   \\
&$\text{WWbL}^{\dagger}$ \cite{shaharabany2022} &$\mathcal{T}$ &VGG16 &27.68 &18.26 &17.37 &19.90 &19.85 &18.70 &21.64 &21.84 &21.75 &21.82   \\  
&\liu{Ours (Step-1)}  &$\mathcal{T}$ &ResNet-50 &33.33 &25.11 &26.47 &23.80 &22.31 &21.61 &22.86 &26.93 &26.62 &27.27  \\
&Ours (Step-2) &$\mathcal{T}$&ResNet-50  &\textbf{44.57} &\textbf{31.17} &\textbf{32.43} &\textbf{29.56} &\textbf{30.90} &\textbf{30.42} &\textbf{30.80} &\textbf{36.00} &\textbf{36.19} &\textbf{36.23}  \\
\hline 
\multirow{7}{*}{PointM} &\liu{AMR}$^\dagger$~\cite{qin2022activation}   &$\mathcal{T}$ &ResNet-50  &7.12 &15.55 &5.52 &28.91 &16.33 &5.90 &30.27 &25.51 &24.96 &26.14   \\ 
&\liu{GroupViT}$^\dagger$ \cite{xu2022groupvit} &$\mathcal{T}$ &GroupViT  &27.44 &25.01 &26.30 &24.42 &25.92 &26.06 &26.12 &30.02 &30.90 &30.98   \\ 
& CLIP-ES$^\dagger$ \cite{lin2022clip} &$\mathcal{T}$ &ViT-Base &52.90 &41.33 &50.61 &30.34 &46.55 &56.20 &33.32 &49.08 &46.22 &45.75 \\ 
&$\text{GbS}^{\dagger}$ \cite{arbelle2021detector} &$\mathcal{T}$  &VGG16 &30.30 &21.58 &19.52 &25.95 &20.95 &18.34 &25.96 &24.64 &24.60 &25.38   \\
&$\text{WWbL}^{\dagger}$ \cite{shaharabany2022} &$\mathcal{T}$ &VGG16  &42.84 &31.28 &31.15 &30.79 &34.47 &33.30 &36.10 &29.32 &32.13 &31.37  \\  
&\liu{Ours (Step-1)} &$\mathcal{T}$  &ResNet-50 &61.70 &51.92 &60.88 &43.02 &40.85 &40.94 &41.13 &52.48 &51.98 &53.29   \\
&Ours (Step-2) &$\mathcal{T}$ &ResNet-50 &\textbf{67.00} &\textbf{54.72} &\textbf{65.64} &\textbf{43.40} &\textbf{53.72} &\textbf{61.30} &\textbf{45.24} &\textbf{58.01} &\textbf{58.84} &\textbf{58.70}   \\

\bottomrule
\end{tabular} 
\vspace{1mm}
\caption{Quantitative comparison \ryn{of different methods using the \textit{IoU} and \textit{PointM} metrics} on four RIS benchmarks.  $Sup.$  denotes the supervision type \ryn{($\mathcal{F}$: full supervision, $\mathcal{B}$: box-level labels, $\mathcal{T}$: text description labels)}. 
(G) and (U) denote the Google and UMD dataset partitions of RefCOCOg. 
\liu{$^\dagger$ indicates the methods adapted from other tasks.}
``-'' denotes \ryn{unavailable values}. 
} 
\label{tab:sota}
\end{table*}

\begin{figure*}[htp]
\begin{center}
\begin{tabular}{c@{\hspace{0.8mm}}c@{\hspace{0.8mm}}c@{\hspace{0.8mm}}c@{\hspace{0.8mm}}c@{\hspace{0.8mm}}c@{\hspace{0.8mm}}c}

\includegraphics[width=0.15\linewidth,height=0.09\linewidth]{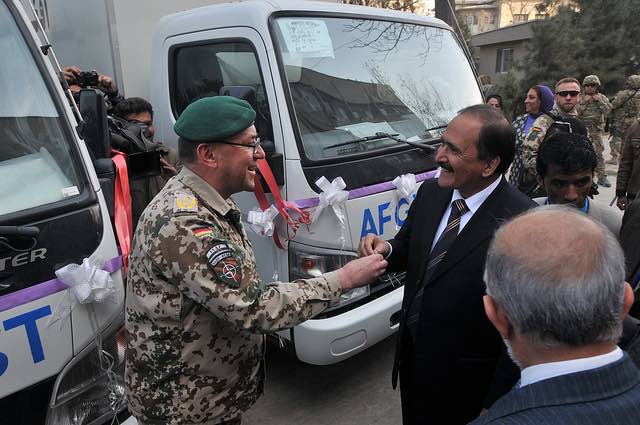}&
\includegraphics[width=0.15\linewidth,height=0.09\linewidth]{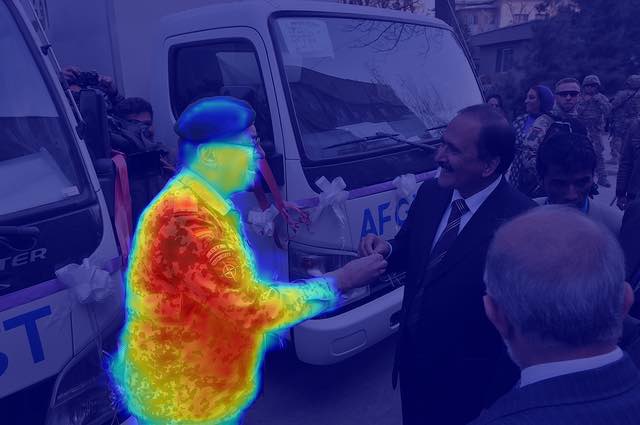}&
\includegraphics[width=0.15\linewidth,height=0.09\linewidth]{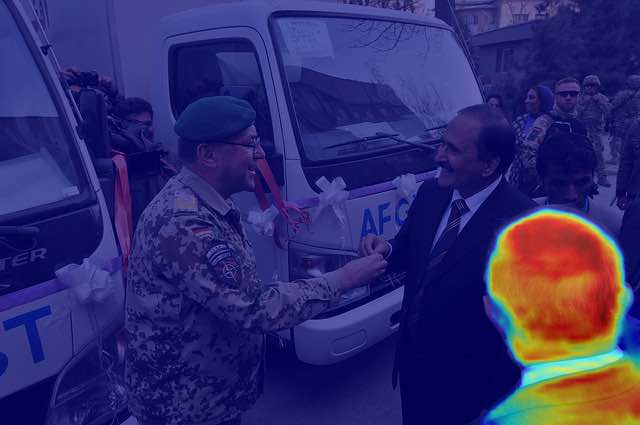}&
\includegraphics[width=0.15\linewidth,height=0.09\linewidth]{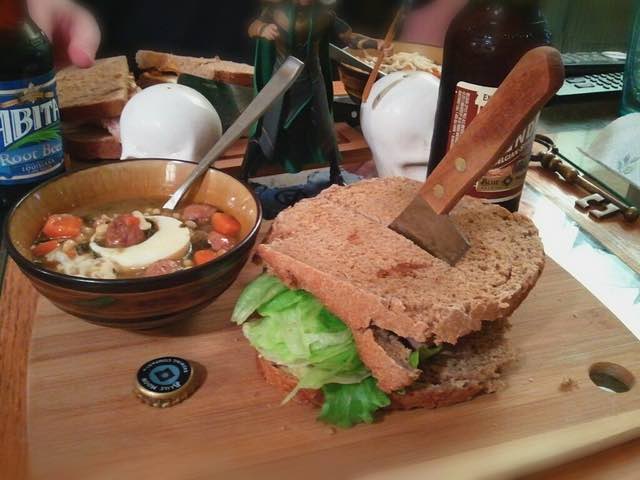}&
\includegraphics[width=0.15\linewidth,height=0.09\linewidth]{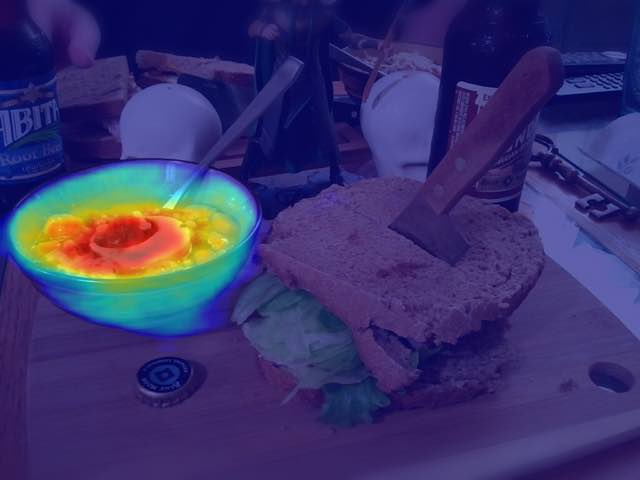}&
\includegraphics[width=0.15\linewidth,height=0.09\linewidth]{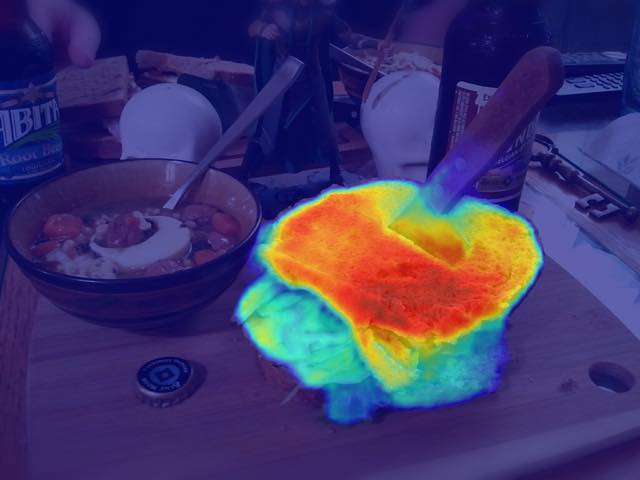}& \vspace{-1mm}\\
\fontsize{8pt}{\baselineskip}\selectfont Image (a)&
\fontsize{7pt}{\baselineskip}\selectfont  \makecell[c]{``man in camouflage  \\ and beret''}&
\fontsize{7pt}{\baselineskip}\selectfont \makecell[c]{``bald head''}&
\fontsize{8pt}{\baselineskip}\selectfont  \makecell[c]{Image (b)}&
\fontsize{7pt}{\baselineskip}\selectfont ``brown bowl of soup''&
\fontsize{7pt}{\baselineskip}\selectfont  \makecell[c]{``a sandwich with \\ 
a knife in it''}&\vspace{1mm}\\
\includegraphics[width=0.15\linewidth,height=0.09\linewidth]{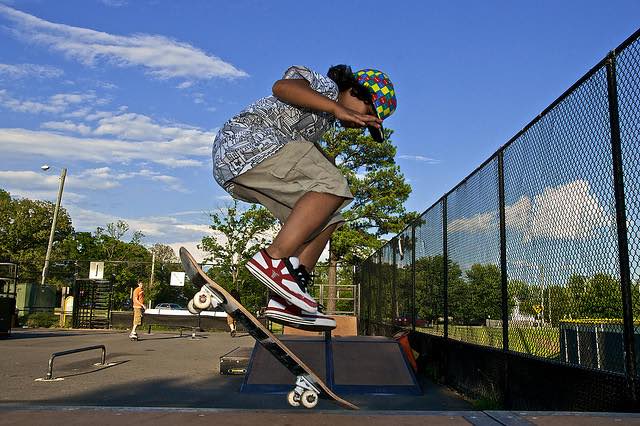}&
\includegraphics[width=0.15\linewidth,height=0.09\linewidth]{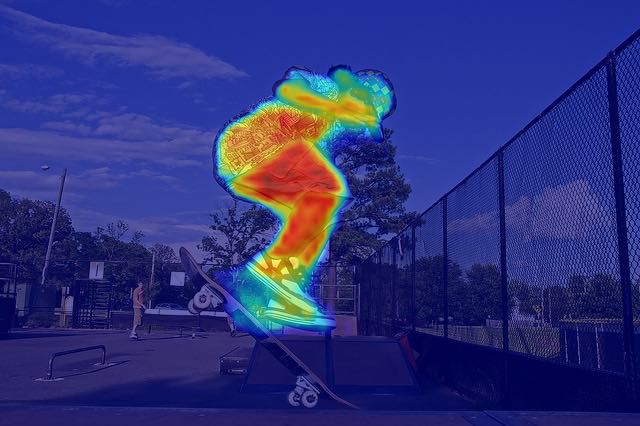}& 
\includegraphics[width=0.15\linewidth,height=0.09\linewidth]{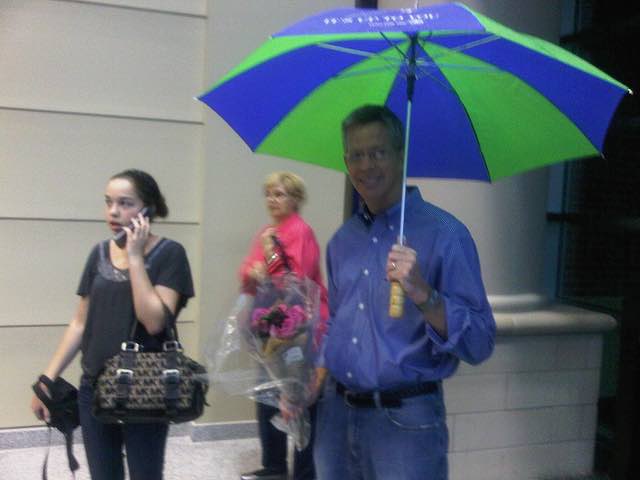}&
\includegraphics[width=0.15\linewidth,height=0.09\linewidth]{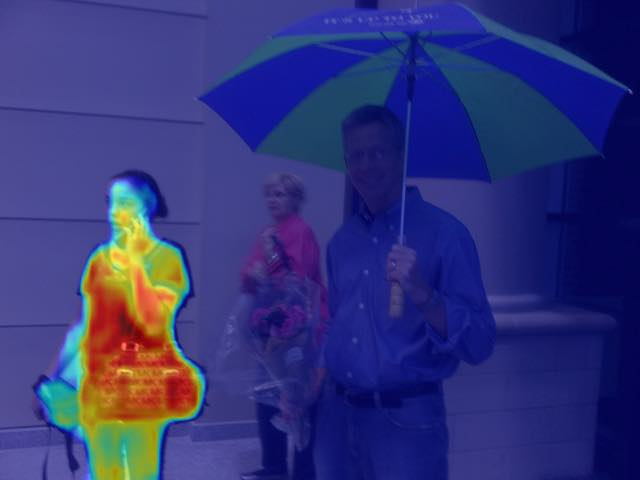}&
\includegraphics[width=0.15\linewidth,height=0.09\linewidth]{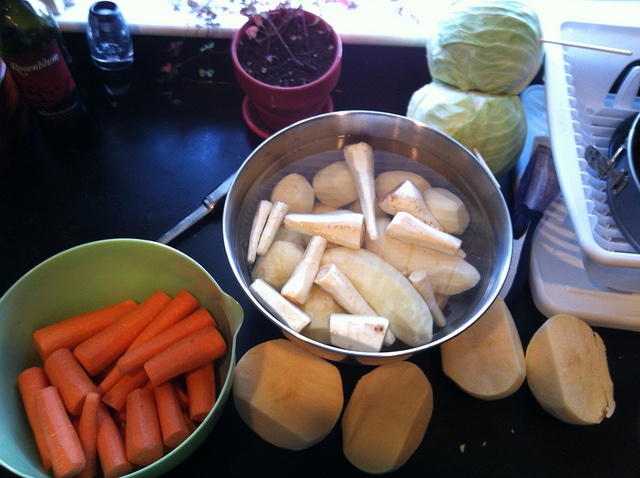}&
\includegraphics[width=0.15\linewidth,height=0.09\linewidth]{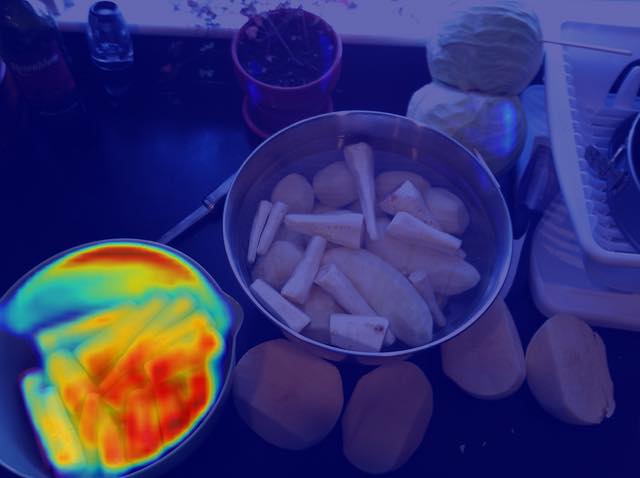}&
\vspace{-1mm}\\
\fontsize{8pt}{\baselineskip}\selectfont Image (c)&
\fontsize{7pt}{\baselineskip}\selectfont  \makecell[c]{``a young boy in a colorful \\ hat jumping a skateboard''}&
\fontsize{8pt}{\baselineskip}\selectfont  \makecell[c]{Image (d)}&
\fontsize{7pt}{\baselineskip}\selectfont \makecell[c]{``a young woman who is \\ standing by a building and \\ talking on a cellphone''}&
\fontsize{8pt}{\baselineskip}\selectfont  \makecell[c]{Image (e)}&
\fontsize{7pt}{\baselineskip}\selectfont  \makecell[c]{``green bowl carrots''}&
\\
\end{tabular}
\end{center}
\vspace{-3mm}
\caption{Qualitative results of \ryn{the proposed RIS} method.}
\label{fig:results}
\end{figure*}

\begin{table}
\begin{center}
\small 
\renewcommand\tabcolsep{1.9pt}
\begin{tabular}{ l | c c c c c c}
    \toprule
Metric & MG \cite{akbari2019multi}   & GbS$^{\dagger}$\cite{arbelle2021detector} &WWbL$^{\dagger}$\cite{shaharabany2022} &\liu{Ours$_{s_1}^*$} &Ours$_{s_1}$ &Ours$_{s_2}$\\
     \midrule
\liu{Acc$_{box}$}  &15.15 &12.67  &24.02 &38.14 &39.90 &\textbf{50.79} \\
PointIt & 47.52  &48.12 &57.04 &70.10 &72.56  &\textbf{74.94}\\
    \bottomrule
\end{tabular}
\end{center}
\vspace{-2mm}
\caption{Quantitative comparison of our method  and  WSGs on the ReferIt \textit{test} set. 
\liu{$^*$ denotes using ImageNet \cite{deng2009imagenet} weights to initialize the image encoder. $s_1$ and $s_2$ denote Step-1 and Step-2, respectively.}
Refer to \ryn{the Supplemental} for more comparisons.
}
\label{tab:sota2}
\end{table}

\subsection{Pseudo Labels Generation and Segmentation}
\label{subsec:step2}
{\flushleft \bf Positive Response Map Selection \fang{(PRMS)}.} There \ryn{are usually} several \ryn{text expressions available} that refer to the same target object in one image, but \ryn{describing different properties of the object}. Although these expressions are discriminative enough for neural networks to localize the target object, the corresponding response maps \ryn{may not be exactly} the same. 
Hence, we select the response map of the best quality by computing the cumulative similarity scores.

Specifically, we first compute the text-to-image response \liu{$\mathbf{R}_{t}^p$} by Eq.~\eqref{eq:response} for each \liu{$\mathbf{Q}_{m}^p$}  of $M$ positive queries.
We then compute the similarity between its masked 
\yu{target} object \yu{with  \ryn{those of all}} positive queries, and sum up all the \yu{object-to-text} similarity scores to reflect the accuracy of the current response map as: 
\begin{align}
   CS_t = \sum_{m=1}^M S(\mathbf{I}, \mathbf{R}_{t}^p, \mathbf{Q}_{m}^p), \ t\in 1,2,\ldots,M.
\end{align}
We select the response map $\mathbf{R}_{t}^p$ with the maximum cumulative score $CS_{t}$ as the response map of the target object.

\yu{{\flushleft \bf Segmentation.}} We use~\cite{ahn2019weakly} to refine our response maps and conduct \ryn{thresholding} on the responses to \ryn{obtain the pseudo labels for training the} segmentation network \ryn{for RIS inference}.
The segmentation network has an image encoder, a text encoder, a multi-modality fusion module, and a decoder. We use the same encoders as in Eq.~\eqref{equ:score}, and the decoder is symmetric to the encoder. 
Since the segmentation network only needs to predict the segmentation map, we use the non-local module~\cite{yang2022lavt} to enrich the visual \ryn{features with the} text features in the last three encoder layers, and send them to the decoder. We train the segmentation network with standard cross-entropy loss ($\mathcal{L}_{ce}$).

\section{Experiments}
\label{sec:experiments}

\subsection{Experiment Setups}
\label{subsec:settings}

{\flushleft \bf Datasets.} 
We conduct experiments on four benchmarks: ReferIt~\cite{kazemzadeh2014referitgame}, RefCOCO~\cite{yu2016modeling}, RefCOCO+~\cite{yu2016modeling} and RefCOCOg~\cite{mao2016generation}. 
ReferIt has $19,894$ images with $13,0525$ annotated referring expressions, which are usually shorter and more succinct. 
The other three datasets are all based on  MSCOCO~\cite{lin2014microsoft}, and each of them contains \ryn{(images, annotated expressions) as: (19,994, 142,209), (19,992, 141,564) and (26,711, 104,560)}. 
While the expressions of RefCOCO focus more on the position property of objects, those of RefCOCO+ focus more on appearance.
Compared to these two datasets, RefCOCOg is more challenging as their expressions \ryn{are usually} longer and more complex. This dataset has two partitions, \ie, the  Google~\cite{mao2016generation} and  UMD~\cite{nagaraja2016modeling} partitions. Both are used in our experiments.

{\flushleft \bf Implementation Details.}
We use ResNet-50 \cite{he2016deep} as our default image encoder, and CLIP \cite{radford2021clip} to initialize the image and text encoders.
Both $\alpha$ and $\beta$ used in our bilateral prompt are set to $0.1$. 
We set the numbers of negative samples in the classification loss and calibration method to $N=47$ and $K=6$, respectively. 
For images that do not have enough negative queries (of only one object with expressions), we randomly sample queries from the rest of dataset to supplement $K$ to $6$.
The number of hidden dimensions is  $C_d=1024$, and the down-sampling ratio is $s=32$.
The loss functions used in Step-1 is $\mathcal{L} = \lambda * \mathcal{L}_{cls} +  \mathcal{L}_{cal}$. $\lambda$ \ryn{is} used to ensure the numerical and gradient equivalence during training, and we empirically set $\lambda$ to $5$.

We implement our framework on PyTorch \cite{paszke2019pytorch} and train it for 15 epochs with a batch size of 48 on an NVIDIA RTX3090 GPU with 24GB of memory. 
During training, we resize the input images to $320\times320$ and set the maximum length of each referring expression to $20$.
The network is optimized using the AdamW optimizer~\cite{loshchilov2017decoupled} with a weight decay of $1e^{-2}$ and an initial learning rate of $5e^{-5}$ with polynomial learning rate decay. The training settings are the same for both steps. \ryn{During} inference, we feed the input image and query text to \ryn{Step-2 of the segmentation network} to produce the segmentation map.

{\flushleft \bf Evaluation Metrics.}
Following \cite{ding2021vln, ye2019cross}, we adopt the mask intersection-over-union (IoU) \liu{and 
Prec@0.5 (P@0.5) metrics} to evaluate the segmentation accuracy.
\liu{Following \cite{akbari2019multi, shaharabany2022}, we also use the pointing-game (PointIt) and box accuracy (Acc$_{box}$) to measure the localization performance.}

However, we note that PointIt tends to yield inaccurate localization scores in our task, as PointIt may count the hit that falls into the box but out of ground truth mask as corrected. Hence, we propose a new metric (PointM), 
\liufang{which is formulated as:
$PointM = \frac{\#Hits}{\#Hits + \#Misses}$, 
where \textit{\#Hits} and \textit{\#Misses} are the numbers of Hits and Misses.
If the maximum point of the response map falls within the ground truth mask regions, it is counted as a Hit. Otherwise, it is considered as a Miss.
}

  \begin{figure*}[t!]
  \centering
  \includegraphics[width=0.9\linewidth]{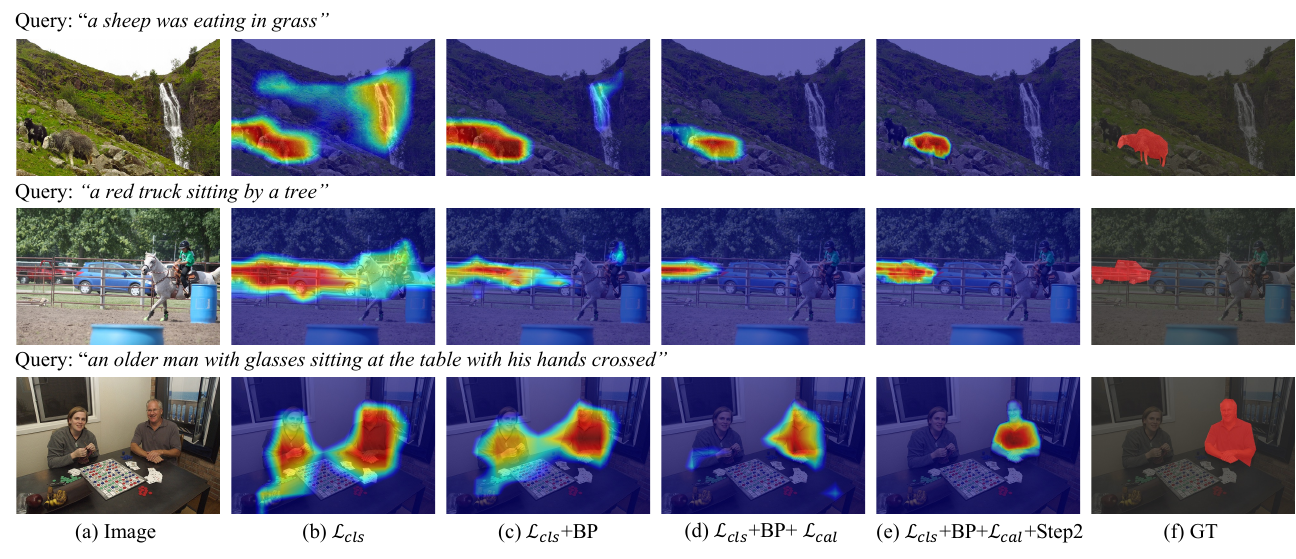}
  \vspace{-3mm} 
  \caption{\ryn{Visualization of the ablation studies to demonstrate} the effectiveness of each component. BP is the bilateral prompt.}
  \label{fig:ablation}
\end{figure*}

\subsection{Comparison to State-of-the-arts}
\label{subsec:comp_sotas}

We validate the effectiveness of our framework by comparing it with fully supervised RIS method 
(LAVT~\cite{yang2022lavt}),
weakly-supervised RIS method (CTS~\cite{feng2022learning}), 
and other related weakly-supervised methods 
(MG~\cite{akbari2019multi},
AMR~\cite{qin2022activation},
GroupViT~\cite{xu2022groupvit},
CLIP-ES~\cite{lin2022clip},
GbS~\cite{arbelle2021detector} and 
WWbL~\cite{shaharabany2022}).

Table~\ref{tab:sota} reports the quantitative comparisons of the segmentation \liufang{and localization} accuracy of different methods on four benchmarks. 
\liu{Our approach demonstrates a significant performance improvement  compared to existing methods \cite{qin2022activation, xu2022groupvit, lin2022clip} that typically generate pseudo-labels using only class labels. 
This is because they lack the capability to discriminate and reason the relations between the instance-level objects in the image, making their adaption to RIS more challenging.
Our framework also exhibits superior performance in comparison to text-based weakly supervised methods \cite{arbelle2021detector, shaharabany2022}.
For instance, it outperforms WWbL with IoU gains of 16.56 and 14.16 on the ReferIt test and RefCOCOg val(G) sets.}
This is due to \liu{three} reasons. First, the formulations of target object localization are different. 
\liu{WWbL uses the output relevance maps of \cite{chefer2021transformer} (which mainly model semantic information of class labels) as GTs to learn response maps. It may be incorrect if \cite{chefer2021transformer} fails to localize the target object.}
In contrast, we locate target objects via the response map generated from text-to-image optimization process, which makes our model more sensitive to information from both modalities.
Second, \liu{our bilateral prompt generates language-guided visual features and image-enhanced text features for better feature alignment.}
\liu{Third}, we leverage instance-wise negative samples to suppress noisy background regions, which calibrates the target's position, while WWbL cannot reduce the background noise.
\liufang{Although CTS achieves better performance than ours, it segments the target object using an auxiliary bounding-box supervision and an offline mask proposal network \cite{arbelaez2014multiscale}. }
In contrast, our approach does not learn from manually annotated masks or bounding boxes. 
The comparison demonstrates that it is possible to train a RIS model only using texts.

In addition, we also report the comparisons of the \liufang{PointIt and Acc$_{box}$} accuracy with WSGs in Table~\ref{tab:sota2}.
On the ReferIt \textit{test} set, our framework \liu{in Step-1} brings PointIt  improvement of 25.04, 24.44, and 15.52 compared to MG, GbS, and WWbL, respectively. 
\liu{Even using pretrained ImageNet \cite{deng2009imagenet} weights to initialize the image encoder, our approach still outperforms WWbL by large margins (Acc$_{box}$: 14.12; PointIt: 13.06).} 
This demonstrates the effectiveness of our framework in locating and segmenting target objects.
Some visual examples are shown in Fig.~\ref{fig:results}, where we can see that our method can localize the target objects in challenging scenes (\eg, \fang{long and complex sentence in Image (d)}) without extra annotations for training. %

\begin{table}
\begin{center}
\small   
\renewcommand\tabcolsep{2pt}
\begin{tabular}{ c  c  c c  c  |c  c c c}
    \toprule
    \multirow{2}{*}{$\mathcal{L}_{cls}$} & \multicolumn{2}{c}{\liu{BP}} & \multicolumn{2}{c|}{$\mathcal{L}_{cal}$}  &\multirow{2}{*}{IoU}&\multirow{2}{*}{PointIt}  &\multirow{2}{*}{\liu{\makecell[c]{P@0.5}}} &\multirow{2}{*}{\liu{PointM}} \\
     &$P_{t2v}$ &$P_{v2t}$  &$P_{term}$&$N_{term}$&  \\
     \midrule
\checkmark &&&&              &18.94 &51.42  &3.87 &36.93 \\ 
\checkmark & \checkmark&&& &19.70 &53.47  &4.21  &39.14 \\ 
\checkmark &\checkmark&\checkmark &                 &        &20.43 &56.49  &4.47  &40.92  \\ 
\checkmark &\checkmark&\checkmark &\checkmark &              &24.18 &64.03 &8.33  &49.85 \\ 
\checkmark &\checkmark&\checkmark & &\checkmark              &22.43 &60.81 &6.44  &45.58 \\ 
\checkmark &\checkmark&\checkmark &\checkmark &\checkmark    &\textbf{27.81} &\textbf{66.99}  &\textbf{12.75}  &\textbf{53.69}\\ 
    \bottomrule
\end{tabular}
\end{center}
\vspace{-2mm}
\caption{Ablation studies of different components on the RefCOCOg (U) \textit{train} set. $P_{t2v}$ and $P_{v2t}$ denote the two variants detached from the proposed bilateral prompt \liu{(BP)}, \yu{which indicate the unilateral prompt from textual to visual features only and the opposite direction. }
$P_{term}$ and $N_{term}$ are the positive enhancement and negative suppression \kk{processes.}} 
\vspace{-4.5mm}
\label{tab:ablation}
\end{table}

\subsection{Ablation Studies and Analyses}
\label{subsec:ablation}
We conduct ablation studies on the text-to-image response map prediction, and report the quantitative results on RefCOCOg (U) train set in Table~\ref{tab:ablation}. 
First, we remove the proposed bilateral prompt and calibration method as our baseline (1-st row), and it can already obtain an initial localization (\eg, PointM: 36.93).
To validate the bilateral prompt, we transform it into two unilateral prompts (\ie, $P_{t2v}$ and $P_{v2t}$) that update only one-way information (\ie, from text to visual or the opposite direction). We can see that the bilateral prompt works better than the unilateral prompts 
and significantly enhances the localization  than the baseline (\eg, PointM increased from 36.93 $\rightarrow$ 40.92).
In addition, we split the calibration method into two processes, \ie, the positive foreground enhancement process ($P_{term}$) and negative background suppression process ($N_{term}$). 
Both improve performance, and when they are combined, performance is further boosted. %

We visualize the effect of each component in Fig.~\ref{fig:ablation}.
Although the model has higher responses at the target object when only Eq.~\eqref{eq:loss_cls} is used, it lacks instance information and suffers from background noise. 
The bilateral prompt enhances the localization of the referred target (see the shades of color in (c)), and attenuates the false responses (\eg, the waterfall in case-1). 
The calibration method can further enhance the correctness by suppressing irrelevant noisy backgrounds (\eg, the black goat in case-1 and the woman in case-3) and maintaining the completeness of the target object.
In (e), Step-2 further improves the performance.

\begin{figure}[t!]
  \centering
   \includegraphics[width=0.9\linewidth]{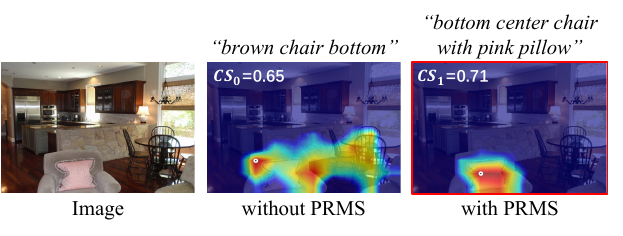}
   \vspace{-3mm}
   \caption{Qualitative comparison of with and without PRMS on RefCOCO+ set. The cumulative score of the response map for each query is placed in the upper left corner of its results. $\odot$ indicates the maximum response point. We highlight the best result with red boxes. 
   }
    \vspace{-2mm}
   \label{fig:same_sents}
\end{figure}

\begin{table}
\begin{center}
\footnotesize  
\renewcommand\tabcolsep{5.5pt}
\begin{tabular}{ l|  c cc c }
    \toprule
     \multirow{2}{*}{IoU}  & \multirow{2}{*}{RefCOCO} & \multirow{2}{*}{RefCOCO+} & \multicolumn{2}{c}{RefCOCOg} \\
     & & & Google & UMD \\
     \midrule
     w/o \liu{PRMS} & 25.11 & 22.31 & 26.93  & 26.62 \\
     w PRMS & \textbf{25.90} &\textbf{ 24.48} & \textbf{27.33}  & \textbf{27.06} \\
    \bottomrule
\end{tabular}
\end{center}
\vspace{-2mm}
\caption{Quantitative comparisons of our method without (w/o) and with (w)  the PRMS on different \textit{val} sets.}
\label{tab:positive}
\end{table}

{\flushleft \bf Positive Response Map Selection (PRMS)}.  
We conduct the add-on experiment on three benchmark\footnote{If the target has only one text expression, \eg, ReferIt dataset, this strategy will degenerate to an identical mapping.} val sets to verify the effectiveness of the positive response map selection strategy.
As shown in Table~\ref{tab:positive}, PRMS generally boosts the performance on all three datasets. 
In particular, it brings around 10\% IoU improvement on the challenging RefCOCO+ dataset. %
In Fig.~\ref{fig:same_sents}, we also show visual comparisons of two cases between the RIS performance with and without PRMS. 
From the comparison, we can see that PRMS is able to select the best response map out of a collection of response maps that corresponds to the queries describing the same object with different properties. This facilitates the training of the segmentation network in Step-2.

{\flushleft \bf Numbers of Negative Queries.}  
The comparison results are shown in Table~\ref{tab:N_K_Cd}. We can see that when we do not use the \ryn{negative text} descriptions for classification, the performance is extremely low.
As the number of negative samples gradually increases, the discriminative power of our model improves and reaches to the peak at $N+1=48$.
When continuing to increase the $N$ to 60, the performance starts to decline instead, due to the huge imbalance of positive and negative queries.
We also investigate the effects of different $K$ in our calibration method, and we can observe a similar trend (the performance is best when $K$=$6$).

\begin{table}
\begin{center}
\footnotesize
\renewcommand\tabcolsep{1.8pt}
\begin{tabular}{ l| c c c c c c c|ccc}
\toprule
\multirow{2}{*}{} & \multicolumn{7}{c|}{$N+1$} & \multicolumn{3}{c}{$K$}  \\
&1 & 2 &6 &12 &24 &48  &60   & 3 & 6 & 12  \\
\midrule
IoU &11.16 &19.91 &22.19 &24.05 &26.19 &\textbf{27.81} &27.61 &27.05 &\textbf{27.81} &27.63  \\
PointIt &16.73 &43.35 &58.18 &64.73 &66.45 &\textbf{66.99} &66.24 &65.81 &\textbf{66.99} &66.82  \\
PointM &8.01 &32.96 &43.71 &49.92 &52.46 &\textbf{53.69} &53.05 &52.94 &\textbf{53.69} &53.16  \\ 
\bottomrule
\end{tabular}
\end{center}
\vspace{-2mm}
\caption{Comparisons of  numbers of negative queries $N$ for  Eq.~\eqref{eq:loss_cls}  and $K$ for  Eq.~\eqref{equ:cali} on the RefCOCOg (U) \textit{train} set. }
\label{tab:N_K_Cd}
\end{table}

\begin{figure}[t!]
  \centering
   \includegraphics[width=1\linewidth]{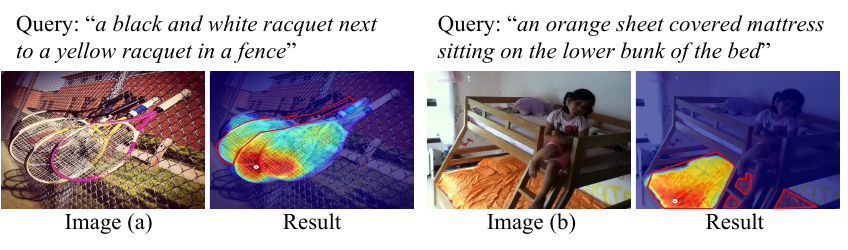}
    \vspace{-6.5mm}
   \caption{Two failure cases. Both images are from the RefCOCOg (U) \textit{val} set, and the results are from Step-2. The GT objects are indicated with red contours for visualization. 
   }
   \vspace{-2mm}
   \label{fig:failure}
\end{figure}

\section{\xk{Conclusion}}
\label{sec:conclusion}
\ryn{In this paper, we have presented a novel RIS framework that uses only the available text descriptions as a supervision signal for training. Our work has three main technical contributions. 
First, we have proposed a bilateral prompt method to help harmonize the discrepancy between the visual and linguistic features. 
Second, we have proposed a calibration method to help reduce background noise to improve the quality of the response maps.
Third, we have proposed a positive response map selection strategy to help obtain high-quality pseudo labels for training a segmentation network for RIS inference.}
\liu{To reduce the in-box error of the existing PointIt metric, we have proposed a new metric (PointM) for a more accurate localization evaluation.}
\yu{Extensive} results demonstrate the effectiveness of our method using only \ryn{text descriptions} as the supervision signal. 

Nonetheless, our \ryn{approach} does have limitations. 
If \ryn{a scene has similar semantics in the foreground and background, it can be distracted by other objects of the same category, and thus produce false localization, \eg, Fig.~\ref{fig:failure}(a)}. 
It \ryn{may also have difficulties in handling  occluded target objects, \eg, Fig.~\ref{fig:failure}(b).}
A possible solution may be to incorporate structured linguistic features with visual features to enhance the model's reasoning abilities.

\noindent \textbf{Acknowledgements:}  This work was supported by the National Key R\&D Program of China \#2018AAA0102003, the Fundamental Research Funds for the Central Universities DUT22JC06, and the National Natural Science Foundation of China 
\#62172073, \#62006037, \#U19B2039 and \#62276046.

{\small
\bibliographystyle{ieee_fullname}
\bibliography{egbib}
}

\end{document}